\PassOptionsToPackage{frozencache=true,cachedir=minted-cache}{minted}
\documentclass[10pt, a4paper]{article}
\usepackage{tipa} 
\usepackage{adjustbox}
\usepackage{tabularx} 
\usepackage{booktabs} 
\usepackage{caption}
\usepackage[final]{lrec2026} 
\usepackage{amsmath}
\usepackage{listings}
\usepackage{xcolor}
\usepackage{multibib}
\usepackage{minted}

\newcites{languageresource}{Language Resources}

\title{A Bolu: A Structured Dataset for the Computational Analysis of Sardinian Improvisational Poetry}

\name{Silvio Calderaro\textsuperscript{1,2}, Johanna Monti\textsuperscript{2}} 

\address{\textsuperscript{1}Università di Pisa, Italia \\
         \textsuperscript{2}Università di Napoli L'Orientale, Italia \\
         silvio.calderaro@phd.unipi.it, jmonti@unior.it\\}

\abstract{
The growing interest of Natural Language Processing (NLP) in minority languages has not yet bridged the gap in the preservation of oral linguistic heritage. In particular, extemporaneous poetry — a performative genre based on real-time improvisation, metrical-rhetorical competence — remains a largely unexplored area of computational linguistics.
This methodological gap necessitates the creation of specific resources to document and analyse the structures of improvised poetry.
This is the context in which A Bolu was created, the first structured corpus of extemporaneous poetry dedicated to\textit{ cantada logudorese}, a variant of the Sardinian language. The dataset comprises 2,835 stanzas for a total of 141,321 tokens. The study presents the architecture of the corpus and applies a multidimensional analysis combining descriptive statistical indices and computational linguistics techniques to map the characteristics of the poetic text.
The results indicate that the production of Sardinian extemporaneous poets is characterised by recurring patterns that support Parry and Lord's theory of formulaicity. This evidence not only provides a new key to understanding oral creativity, but also offers a significant contribution to the development of NLP tools that are more inclusive and sensitive to the specificities of less widely spoken languages.
 \\ \newline \Keywords{ Corpus of minority languages, Extemporaneous poetry, Oral-formulaic patterns} }

\begin{document}

\maketitleabstract

\section{Introduction and Background}

The growing interest of the Natural Language Processing (NLP) community in low-resource languages, minority varieties and dialects reflects a broader shift toward linguistic inclusivity. For decades, computational tools and annotated resources were concentrated on a small set of high-resource languages, leaving the vast majority of the world's linguistic heritage outside the reach of modern NLP techniques. Corpus construction efforts for minority languages must typically contend with sparse digital presence, orthographic instability, the absence of standardized annotation frameworks, challenges well attested in comparable projects for languages such as Guarani~\cite{guarani2022}, Breton~\cite{arbres2024} and the minority languages of Italy~\cite{ramponi2022}. These difficulties are further compounded when the target variety is rooted in oral and performative traditions \cite{zumthor1997construction, ong1982orality} that have historically resisted stabilization into annotated digital formats.

Within this landscape, Sardinian occupies a particularly complex position. Officially recognized as a minority language under Italian national law and widely regarded as the closest living descendant of Latin~\cite{virdis1988}. Existing resources include SardNet~\cite{angioni2018}, a lexical resource mapping Sardinian word senses onto WordNet entries, a BERT-based Part-of-Speech tagger~\cite{carta2025} and a nascent Automatic Speech Recognition system~\cite{chizzoni2024}. 

The \textit{cantada logudoresa} is a formal poetic contest in which \textit{cantadores} improvise verses in the Logudorese dialect under strict metrical and thematic constraints, structured around the \textit{Otada}, an octave of eight hendecasyllabic lines following an ABABABCC or ABBABACC rhyme scheme, alongside shorter forms such as the \textit{batorina} and the closing \textit{dispedida}. Performers must simultaneously manage metrical and rhyming complexity, thematic coherence,  real-time argumentative exchange and all under the pressure of live performance. The preservation of this tradition has historically depended on inconsistent and scattered acts of transcription, precluding any systematic computational investigation. This paper introduces \textbf{A Bolu} \citeplanguageresource{ABolu2026}, the first structured digital corpus of the \textit{cantada logudoresa}, designed to fill this gap and establish a reproducible framework for its quantitative and linguistic analysis.
Our contributions:
\begin{enumerate}
    \item \textbf{Digital Preservation and Resource Creation:} We provide a high-fidelity digital repository for a vulnerable minority language tradition, preventing the loss of undocumented or fragmented transcriptions and establishing a foundation for future NLP tasks in Sardinian.
    \item \textbf{Multidimensional Data Modeling:} Unlike flat-text corpora, \textit{A Bolu} is structured to include rich metadata—such as thematic assignments, performer identifiers and precise execution timestamps per stanza—modeled in a hierarchical format to facilitate complex relational queries.
    \item \textbf{Computational Stylistics Analysis:} We demonstrate the utility of the dataset by proposing it as a benchmark for investigating "stylistic signatures" and lexical complexity, enabling quantitative research into how real-time improvisational pressures affect the linguistic and metrical choices of the \textit{cantadores}.
\end{enumerate}
The aim of this resource is to integrate Sardinian extemporaneous poetry into the field of computational linguistics, defining a reproducible framework that promotes new avenues of study and formal analysis of structured oral traditions.

\section{Methodology}
This section describes the methodological approach adopted for the construction of the corpus, divided into the phases of data acquisition \ref{data_acquisition}, archive structuring \ref{data_structure} and curation of the raw material \ref{data_processing}. The central objective was to transform a heritage of oral tradition—fragmentary and discontinuous by nature—into a structured digital resource, suitable for computational processing and quantitative linguistic analysis.

\subsection{Data Acquisition and Source Selection} \label{data_acquisition}
The primary data for this study were collected programmatically from làcanas.it \cite{lacanas_poetasabolu}, an online newspaper dedicated to Sardinian culture, news, identity, with an archive of improvised poems in the Sardinian language. The scraping process targeted the poems on the site to maintain the same origin as the source. The corpus was constructed according to strict criteria of linguistic and typological homogeneity to ensure the reliability of subsequent quantitative comparisons. The sources were selected based on three fundamental parameters:

\begin{enumerate}
    \item \textbf{Transcription Quality and Metadata Richness:}produced by the official editorial staff were included in the corpus. This restriction was implemented to ensure consistency, reliability and philological accuracy across the dataset. A primary selection criterion was the availability of essential contextual metadata, including the performance setting, the designated themes (i.e., thematic debates), and  the recorded execution time of each stanza for each poet.
    \item \textbf{Generic Consistency:} To avoid stylistic bias, the dataset exclusively comprises performances belonging to the same poetic genre, specifically\textit{(cantada logudoresa)}. This ensures that the metrical constraints and the thematic development remain constant across the entire sample.
    \item \textbf{Linguistic Variety:} The selection was restricted to a single linguistic variety of the Sardinian language (\textit{Logudorese}). This choice eliminates lexical variation due to dialectal shifts, allowing the analysis to focus strictly on the individual poets' lexical complexity and rhyming strategies.
\end{enumerate}
Despite the application of these selection criteria, the corpus inevitably reflects the intrinsic challenges associated with documenting oral traditions. It should be noted that several performances within the dataset are incomplete: in some cases, individual stanzas are partially  or entirely absent from the original transcriptions. These lacunae—often resulting from recording interruptions or archival deterioration—were systematically identified during the data-cleaning phase in order to prevent distortions in the statistical distribution of the linguistic metrics.
Furthermore, the texts included in the corpus do not represent complete poetry contests. Rather, the materials primarily consist of the central phase of the competition, which accounts for the majority of the recorded performances. Consequently, these contest-based transcriptions frequently omit either the opening segment (\textit{esordiu)} or the final closing exchange (\textit{dispèdida}) of the debate.

\subsection{Data Structure and Corpus Architecture} \label{data_structure}

The collected performances were encoded in a structured, hierarchical format using \texttt{JSON} (JavaScript Object Notation). This data model was designed to represent the multi-layered organisation of extemporaneous poetic debates, preserving the internal segmentation of each performance while maintaining explicit links between individual stanzas and their associated metadata. Such a structure ensures both formal consistency and computational accessibility, thereby facilitating corpus querying, annotation and quantitative analysis.

The corpus is organized into two primary levels:

\begin{itemize}
    \item \textbf{Global Metadata:} This top-level object captures the contextual framework of the entire performance. It includes the \textit{title} of the debate, the \textit{source URL} for traceability, an \textit{introductory summary} of the event and the \textit{central theme} (thematic dispute) assigned to the poets.
    \item \textbf{Transcription Units (Stanzas):} The \texttt{transcription} field contains an ordered array of stanza-level objects, each corresponding to a single metrical unit: namely an octave (\textit{Otada}), quatrain (\textit{Batorina}), couplet (\textit{Duina}), or closing/free-form stanza (\textit{Dispèdida}). Each object encodes a discrete poetic turn and is associated with a set of structured attributes: a unique numerical \texttt{id} (ensuring sequential traceability), the \texttt{poet} identifier, the \texttt{metrics} label specifying the metrical form, the recorded \texttt{time} of execution and an ordered list of \texttt{verse} strings representing the individual lines of the stanza.
\end{itemize}

An example of the data representation for a single stanza and its associated metadata is provided below:

\begin{minted}[
    frame=none,
    breaklines=true,
    fontsize=\footnotesize
]{json}
{
  "metadata": {
    "title": "Sozu e Masala in Ballao: tempus e omine",
    "intro_theme": "Cando cantaian manu-manu Peppe Sozu e Marieddu Masala sa resultada de sa gara fut bella e assigurada in partenzia. [...]",
    "core_theme": "Disputa tra l'Eternita inarrestabile e l'Ingegno Creativo",
    "source": "https://www.lacanas.it/..."
  },
  "transcription": [
    {
      "id": 1,
      "poet": "Sozu",
      "metrics": "Otada",
      "time": "1'18",
      "verse": [
        "Ja non cherio in piata sa zente",
        "chi da s'atesa restet in fastizu.",
        "Como sighimos che babbu e che fizu",
        "tantu de allegrare s'ambiente",
        "ja chi su comitadu intelligente",
        "at apagadu su nostru disizu",
        "ch'in parte 'e unu tema mi collocat",
        "e a cantare su tempus mi tocat."
      ]
    }
  ]
}
\end{minted}
This hierarchical architecture transforms the poetic performance from a static text into a multidimensional data object. By explicitly linking linguistic output to specific constraints—such as the thematic assignment, the metric of the stanzas and the execution time—the dataset allows for a granular investigation. Furthermore, this structured format ensures the corpus is fully interoperable with modern NLP pipelines, establishing a reproducible framework for the quantitative study of Sardinian oral traditions.

\subsection{Data Processing and Corpus Curation} \label{data_processing}

Once the raw data were extracted, a rigorous process of curation and normalization was required to transform the scraped material into a reliable research dataset. This phase involved both automated filtering and manual philological verification to address the inconsistencies inherent in a heterogeneous digital archive.

\begin{enumerate}
    \item \textbf{Deduplication Record :} A primary challenge was the presence of duplicate performances. Many poetic debates were found to be published multiple times under slightly different titles or categorized in different sections of the source website. These redundant entries were identified and removed to ensure that the statistical analysis of lexical frequency and poet participation remained unbiased.
    \item \textbf{Entity Resolution and Normalization:} To ensure that each poet's stylistic signature was correctly attributed, we performed a normalization of personal names. Variations in transcription, such as the inconsistent use of accents (e.g., \textit{Màsala} vs. \textit{Masala}), were reconciled to a single canonical form. This step is crucial for the subsequent calculation of individual lexical complexity and comparative stylometry.
    \item \textbf{Structural Integrity and Lacunae Flagging:} Each stanza was checked automatically and manually to verify its completeness. Given the oral and often fragmented nature of the transcriptions, we adopted a symbolic tagging system within the \textit{metrical form} metadata field to maintain the chronological sequence of the debate without compromising the linguistic statistics:
    \begin{itemize}
        \item \textbf{Standard labels} (e.g., \texttt{otada}): Applied to complete stanzas where all verses are present.
        \item \textbf{Single asterisk} (e.g., \texttt{otada*}): Indicates a partially missing stanza where only a portion of the verses was transcribed.
        \item \textbf{Double asterisk} (e.g., \texttt{otada**}): Indicates a fully missing stanza, preserving its original position in the performance flow while excluding it from the textual analysis.
    \end{itemize}
    \item \textbf{Temporal Standardization:} The execution time for each stanza, recorded in the source as a string format (e.g., \texttt{1'00''}), was parsed and converted into a discrete numerical variable representing total seconds. In cases where the timing was not present in the original source, or when the stanza was incomplete (as indicated by the asterisk system), a \texttt{null} value was assigned to the field. 
\end{enumerate}

This approach ensures that the resulting corpus is not merely a collection of texts, but a structured digital resource that preserves both the textual content and the contextual information of the poetic contests, including stanza-level metadata, performance timing and thematic annotation.

\section{Corpus statistics}
The resulting dataset, \textbf{A Bolu}, to the best of our knowledge constitutes the first structured digital corpus of Sardinian extemporaneous poetry explicitly designed for computational analysis. By aggregating dispersed transcriptions and imposing a systematic structural organization, the corpus provides a solid foundation for rigorous empirical investigation not only into this minority language tradition but also into the structure and dynamics of extemporaneous poetic performance itself.

\subsection{Statistical overview} \label{Statistical_overview}

The final corpus, consists of 55 digitalized poetic sessions with a specific focus on the 20th-century oral heritage. 
Table~\ref{tab:general-stats} presents a synoptic view of the corpus dimensions. The dataset comprises 2,835 stanzas distributed across 55 poems, yielding an average of 51.55 stanzas per poem. The corpus also documents its own gaps transparently: 60 stanzas lack execution time annotations, 63 are partially incomplete in their textual transcription and 8 are entirely missing. Rather than discarding these entries, they have been retained with explicit flags to preserve the structural integrity of each poetic debate.

\begin{table}[h]
\centering
\begin{tabular}{lr}
\hline
\textbf{Feature} & \textbf{Value} \\ \hline
Total number of poems (JSON files) & 55 \\
Total number of stanzas & 2,835 \\
Average stanzas per poem & 51.55 \\
Unique identified poets & 8 \\
Stanzas with missing execution time & 60 \\
Stanzas with partially incomplete & 63 \\
Stanzas with totally incomplete & 8 \\ \hline
\end{tabular}
\caption{General statistics of the A Bolu corpus.}
\label{tab:general-stats}
\end{table}

\subsection{Poet contributions and metrical distribution}
The following section examines how stanzas are distributed across poets 
and metrical forms within the corpus, highlighting both the dominance of certain performers and the structural variety of the poetic debate. Table~\ref{tab:poets-detailed-granular} provides a comprehensive breakdown of stanza counts per poet, disaggregated by metrical type and transcription completeness.
Two figures dominate the dataset: Sozu and Masala each appear in 32 poems and contribute 811 and 753 stanzas respectively, together accounting for over 55\% of the total corpus. This concentration reflects their historical prominence within the tradition and ensures substantial material for in-depth stylistic analysis. Piras (493 stanzas, 20 poems) Mura (339 stanzas, 15 poems) and Piredda (246 stanzas, 10 poems) constitute a second tier, while the remaining three poets, Sale, Seu, and Budrone, contribute smaller but nonetheless valuable samples, particularly for contrastive purposes.
Regarding metrical variety, the \textit{Otada} is overwhelmingly the foundational unit of the corpus, comprising 2,592 of the 2,835 stanzas (93.9\%). The remaining forms,\textit{Duina} (95 stanzas, 3.4\%), \textit{Batorina} (70 stanzas, 2.5\%), \textit{Despedida} (7 stanzas, 0.2\%), appear with considerably lower frequency, consistent with their specialized roles within the structure of the poetic debate. The distribution of these secondary forms is not uniform across the poets.
\textit{Masala} employs the widest metrical repertoire, featuring 49 \textit{Duinas},
35 \textit{Batorinas}, and 3 \textit{Despedidas}. This is followed by \textit{Sozu} (20 \textit{Duinas}, 16 \textit{Batorinas}, 2 \textit{Despedidas}), \textit{Mura} (16 \textit{Duinas}, 7 \textit{Batorinas}, 1 \textit{Despedida}),  \textit{Seu} (10 \textit{Duinas}, 12 \textit{Batorinas} and 1 \textit{Despedida}). In contrast, \textit{Piras}, \textit{Piredda}, \textit{Sale} and \textit{Budrone} perform exclusively in the \textit{Otada} form. This asymmetry is likely attributable to the uneven distribution of available data across poets in the original source.
The specifically tagged entries \textit{Otada*} (63 stanzas) and \textit{Otada**} (8 stanzas) denote partially and totally incomplete transcriptions, respectively. These lacunae are distributed unevenly across poets: Sozu accounts for the largest share of partial gaps (21 \textit{Otada*}), a figure that is proportional to his extensive presence in the corpus and likely reflects the variable quality of the source recordings. By retaining these entries with explicit markers rather than omitting them, the dataset preserves the sequential structure and turn-taking logic of each session, which is essential for any analysis of debate dynamics and interactional patterns between competing poets.

\begin{table*}[t]
\centering
\begin{tabular}{l|r|rrr|rr|r|r}
\hline
\textbf{Poet} & \textbf{Tot.} & \textbf{Otada} & \textbf{Otada*} & 
\textbf{Otada**} & \textbf{Duina} & \textbf{Bator.} & \textbf{Disped.} 
& \textbf{Poems} \\ \hline
Sozu    & 811 & 751 & 21 & 1 & 20 & 16 & 2 & 32 \\
Masala  & 753 & 656 &  6 & 4 & 49 & 35 & 3 & 32 \\
Piras   & 493 & 472 & 18 & 3 &  0 &  0 & 0 & 20 \\
Mura    & 339 & 305 & 10 & 0 & 16 &  7 & 1 & 15 \\
Piredda & 246 & 242 &  4 & 0 &  0 &  0 & 0 & 10 \\
Sale    &  88 &  85 &  3 & 0 &  0 &  0 & 0 &  4 \\
Seu     &  86 &  62 &  1 & 0 & 10 & 12 & 1 &  4 \\
Budrone &  19 &  19 &  0 & 0 &  0 &  0 & 0 &  1 \\ \hline
\textbf{Total} & \textbf{2,835} & \textbf{2,592} & \textbf{63} & 
\textbf{8} & \textbf{95} & \textbf{70} & \textbf{7} & \textbf{-} \\ 
\hline
\end{tabular}
\caption{Comprehensive breakdown of stanzas per poet, including metrical 
forms and transcription lacunae (\textit{Otada*} = partially incomplete; 
\textit{Otada**} = totally incomplete).}
\label{tab:poets-detailed-granular}
\end{table*}

\subsection{Lexical analysis}

In the context of oral improvised poetry, the lexical dimension acquires particular significance: poets must generate verse in real time under strict metrical and rhyming constraints, drawing on a mental lexicon that is simultaneously broad enough to avoid repetition and sufficiently controlled to satisfy the formal requirements of the tradition. Measuring lexical richness in such a setting can shed light on the cognitive and linguistic resources that differentiate one performer from another.
To ensure a methodologically robust evaluation and to mitigate the well-documented sensitivity of the standard Type-Token Ratio (TTR) to corpus length \cite{baayen2001word}, we employ two length-independent indices: the Moving Average Type-Token Ratio (MATTR) \cite{covington2010cutting} and the Measure of Textual Lexical Diversity (MTLD) \cite{mccarthy2010mtld}. For the MATTR calculation, a sliding context window of 50 words was selected, as it approximates the average length of a single \textit{ottava}, the fundamental metrical unit of the performance, thus capturing lexical density at the level of individual poetic turns.

Table~\ref{tab:lexical-stats} reports token counts, type counts, hapax legomena and the calculated lexical diversity indices for each poet. The full corpus comprises 141,321 tokens and 12,973 types, yielding an overall TTR of 9.18\%. This low figure is consistent with the lexical saturation expected in a genre-constrained oral corpus, where recurrent grammatical forms and thematic vocabulary inevitably accumulate across large text volumes. Individual sub-corpora range from 968 tokens (Budrone) to 41,164 tokens (Sozu), a difference of more than fortyfold, which makes direct TTR comparisons across poets methodologically unreliable.

As expected, raw TTR values vary considerably, from 14.47\% (Sozu) to 44.01\% (Budrone), a divergence that reflects corpus size rather than genuine differences in lexical competence. When examining the MATTR (window = 50), however, scores are remarkably stable across all eight performers, with values ranging from 79.98 (Budrone) to 83.26 (Mura) and a mean of $\mu = 81.59$. This convergence suggests that local lexical density -- measured within windows that approximate the length of a single \textit{ottava} -- constitutes a structural property of the \textit{Cantada a bolu} tradition, largely independent of the total volume of text produced by each poet.


The MTLD scores provide a complementary perspective on sustained lexical diversity. The global MTLD computed on the entire concatenated corpus is 95.87, with individual values ranging from 81.20 (Budrone) to 110.77 (Mura). These figures indicate a consistent capacity for lexical variation throughout extended stretches of discourse. Mura's score of 110.77 stands out as the highest in the sample, suggesting an exceptional ability to vary vocabulary over longer sequences, even when compared to poets with larger sub-corpora such as Sozu or Masala.

The hapax legomena counts offer a further dimension of analysis. Across the full corpus, 6,789 word forms occur exactly once, accounting for approximately 52.33\% of the 12,973 distinct types attested. This substantial proportion of singletons suggests that the poets do not rely exclusively on a fixed repertoire of formulaic expressions but instead introduce novel vocabulary throughout their performances. Taken together, the MATTR and MTLD results indicate that the \textit{A Bolu} corpus exhibits a high and consistent level of lexical richness across sub-corpora of considerably different sizes, lending support to the view that local lexical density is a stable feature of this oral poetic tradition.

\begin{table}[h]
\centering
\small
\begin{adjustbox}{max width=\columnwidth}
\begin{tabular}{lrrrrrrr}
\hline
\textbf{Author} & \textbf{Tokens} & \textbf{Types} & \textbf{Hapax} & \textbf{TTR\%} & \textbf{MATTR} & \textbf{MTLD} \\ \hline
Sozu    &  41,164 & 5,956 & 3,295 & 14.47 & 81.89 &  97.13 \\
Masala  &  36,305 & 5,779 & 3,411 & 15.92 & 81.80 &  98.82 \\
Piras   &  25,573 & 4,435 & 2,712 & 17.34 & 80.56 &  85.65 \\
Mura    &  16,507 & 3,302 & 2,053 & 20.00 & 83.26 & 110.77 \\
Piredda &  12,440 & 2,701 & 1,674 & 21.71 & 80.79 &  94.85 \\
Sale    &   4,576 & 1,344 &   915 & 29.37 & 82.60 & 100.10 \\
Seu     &   3,788 & 1,226 &   825 & 32.37 & 82.40 & 104.93 \\
Budrone &     968 &   426 &   301 & 44.01 & 79.98 &  81.20 \\ \hline
\textbf{Total} & \textbf{141,321} & \textbf{12,973} & \textbf{6,789} & \textbf{9.18} & \textbf{81.59} & \textbf{95.87} \\ \hline
\end{tabular}
\end{adjustbox}
\caption{Lexical statistics per author and corpus totals: tokens, types, hapax legomena, TTR, MATTR and MTLD.}
\label{tab:lexical-stats}
\end{table}

\subsection{Temporal Dynamics and Metric Complexity}

The inclusion of execution time as a granular attribute introduces an important dimension for analysis in this specific domain. Table \ref{tab:primary-execution-times} summarizes the average duration (in seconds) for the three primary metrical forms identified: the \textit{Otada}, the \textit{Duina} and the \textit{Batorina}.

The data reveal a clear correlation between metrical length and temporal duration. The \textit{Otada}, being the foundational and most complex unit of the \textit{cantada}, requires a corpus-wide average of 58.18 seconds per unit. In contrast, the \textit{Duina} and the \textit{Batorina} function as more rapid improvisational units,due to the brevity of their short nature, with average execution times of 10.02 and 24.93 seconds, respectively.

An analysis of individual poet performances highlights significant variations in improvisational pace:
\begin{itemize}
    \item \textbf{Masala} emerges as the most rapid improviser across all categories, with an average \textit{Otada} time of 45.57 seconds, significantly below the general mean of 58.18 seconds.
    \item \textbf{Sozu}, despite being one of the most prolific contributors, maintains a more measured pace with an average of 68.33 seconds per \textit{Otada}.
    \item \textbf{Sale} recorded the highest average time for a complete octave (83.25 seconds), suggesting a different stylistic approach to the rhythmic constraints of the performance.
\end{itemize}

It should be noted that these poetic performances are mainly sung and often accompanied by musical interludes. For this reason, it would be misleading to correlate temporal variations with lexical data, as this would not provide truly meaningful information about the creative abilities of the poets.
Some poets, in fact, prefer to dwell on the vowels /\textipa{a}/, /\textipa{e}/ and /\textipa{\textepsilon/}/.

For this reason, temporal characteristics should be considered as stylistic traits specific to each poet, rather than as descriptors of actual improvisational abilities.

\begin{table}[h]
\centering
\small
\setlength{\tabcolsep}{4pt} 
\begin{tabular}{lrrr}
\hline
\textbf{Author} & \textbf{Otada (s)} & \textbf{Duina (s)} & \textbf{Batorina (s)} \\ 
\hline
Sozu & 68.33 & 11.55 & 26.69 \\
Masala & 45.57 & 8.51 & 20.80 \\
Piras & 53.83 & -- & -- \\
Mura & 65.41 & 12.00 & 31.71 \\
Seu & 66.95 & 10.90 & 30.67 \\
Sale & 83.25 & -- & -- \\
Piredda & 49.20 & -- & -- \\
Budrone & 56.89 & -- & -- \\
\hline
\textbf{Mean (All)} & \textbf{58.18} & \textbf{10.02} & \textbf{24.93} \\
\hline
\end{tabular}
\caption{Average execution time (seconds) for the primary metrical forms per author.}
\label{tab:primary-execution-times}
\end{table}

\begin{table*}[h]
\centering
\footnotesize
\renewcommand{\arraystretch}{1.2}
\setlength{\tabcolsep}{3pt}
\begin{tabular}{p{4.8cm}p{2.3cm}ccccp{2.8cm}}
\hline
\textbf{Verse / Template} & \textbf{Poet(s)} & \textbf{$n$} & \textbf{Obs.} & \textbf{PMI} & \textbf{LLR} ($G^2$) & \textbf{Type} \\
\hline
\multicolumn{7}{l}{\textit{FC 1 — Deictic formulas}} \\
\hline
\textit{de fronte a unu pópulu [\textit{ADJ}]}
    \newline \scriptsize(transl.\ \textit{in front of a [ADJ] people})
    & Masala        & 5 & 8 & 27.29 & 286.68 & Personal deixis formula \\
\textit{de su palcu in susu}
    \newline \scriptsize(transl.\ \textit{from the stage up above})
    & Mura / Masala & 5 & 4 & 22.95 & 119.27 & Shared deixis formula \\
\hline
\multicolumn{7}{l}{\textit{FC 2 — Personal turn-management formulas}} \\
\hline
\textit{[\textit{ADV/PRON}] una cosa narrer ti cheria}
    \newline \scriptsize(transl.\ \textit{[ADV/PRON] one thing I would like to tell you})
    & Sozu          & 5 & 5 & 29.44 & 194.09 & Personal formula (var.\ onset) \\
\textit{su chi ses nelzende pone cura}
    \newline \scriptsize(transl.\ \textit{pay attention to what you are (weaving) saying})
    & Piredda       & 6 & 3 & 42.72 & 171.66 & Personal formula \\
\hline
\multicolumn{7}{l}{\textit{FC 3 — Dialogic mirroring}} \\
\hline
\textit{a morrer de piumu est un'onore}
    \newline \scriptsize(transl.\ \textit{to die by lead (a bullet) is an honour})
    & Piras $\to$ Sozu  & 8 & 2 & 46.62 & 125.25 & Dialogic mirroring \\
\textit{chi si l'at mandigada sa rustia}
    \newline \scriptsize(transl.\ \textit{that the blight has devoured it})
    & Masala $\to$ Mura & 7 & 2 & 45.50 & 122.15 & Dialogic mirroring \\
\hline
\multicolumn{7}{l}{\textit{FC 4 — Formulaic variants \& cross-poet templates}} \\
\hline
\textit{e cando mai chi non b'as cumpresu}
    \newline \scriptsize(transl.\ \textit{and when have you ever not understood it})
    & Piredda       & 8 & 2 & 45.80 & 122.98 & Cross-poet template \\
\textit{e cando mai chi non l'as cumpresa}
    \newline \scriptsize(transl.\ \textit{and when have you ever not understood it})
    & Sozu          & 8 & 2 & 45.57 & 122.34 & Cross-poet template \\
\textit{[\textit{V}] est chi non l'as [\textit{V-PART}]}
    \newline \scriptsize(transl.\ \textit{[it] is that you have not [V-PART] it})
    & Mura / Sale   & 5 & 3 & 16.71 &  63.48 & Cross-poet template \\
\hline
\end{tabular}
\caption{Recurring $n$-grams ($n = 5$--$8$) grouped by functional
category (FC), ordered by LLR within each group.
\textit{Obs.}\ = observed frequency;
PMI = Pointwise Mutual Information ($\log_2$);
LLR = Log-Likelihood Ratio ($G^2$), used as a relative ranking measure;
no significance thresholds are applied.
Square brackets denote variable slots.}
\label{tab:ngram_functional_categories}
\end{table*}

\subsection{Co-occurrence and Formularity}

To investigate recurrent multiword patterns in the corpus, we conducted
an exhaustive $n$-gram analysis with $n$ ranging from 3 to 8. 
The analysis is exploratory in nature: its aim is not to establish statistically validated categories of formulaic reuse, but to identify recurring multiword patterns and examine their functional distribution across poets and performance events.
The analysis was carried out on the full transcribed corpus of
135,747 tokens and 12,808 unique word forms, comprising
performances exclusively in the \textit{ottava} meter (\textit{otada}).
Token extraction followed a conservative normalization pipeline: lowercase conversion, removal of punctuation while preserving apostrophized clitics, whitespace normalization, with original surface forms retained separately for display purposes.
Shorter sequences ($n \leq 4$) are more likely to include 
high-frequency grammatical collocations and are therefore less 
diagnostically reliable in isolation; sequences of five tokens 
or more, especially when recurring across distinct performance 
events or poets, provide stronger evidence of formulaic reuse, 
with higher-order sequences ($n \geq 6$) offering the most 
robust signal, being least likely to occur by chance.

For each candidate sequence, two association measures were computed.
Pointwise Mutual Information~\cite{church1990} is defined as:
\[
\text{PMI} = \log_2 \frac{P(w_1, \ldots, w_n)}{\prod_{i=1}^{n} P(w_i)}
\]
where $P(w_1, \ldots, w_n)$ is the observed joint probability of the
sequence and $\prod P(w_i)$ is the expected probability under lexical
independence.
PMI is a reliable indicator of lexical \emph{distinctiveness} but
unstable at low observed frequencies, as sequences containing rare items yield inflated values.
It was therefore supplemented with the Log-Likelihood
Ratio~\cite{dunning1993}, $G^2$:
\[
G^2 = 2 \sum_{i} O_i \ln \frac{O_i}{E_i}
\]
where $O_i$ and $E_i$ are observed and expected cell counts contrasting occurrences of the sequence against all other corpus positions.
$G^2$ is more stable at low frequencies and penalizes sequences whose association score is driven by constituent rarity rather than genuine co-occurrence preference.
Since all $n$-gram candidates were evaluated within a single 
exploratory pass over the corpus, $G^2$ values are not 
interpreted as hypothesis tests: multiple comparison corrections 
would be required for inferential validity and given the 
exploratory nature of the analysis we adopt $G^2$ strictly as a \emph{relative ranking measure} of lexical cohesion.

The verses considered are reported in Table~\ref{tab:ngram_functional_categories}, grouped into four functional categories assigned on the basis of qualitative distributional and pragmatic criteria, reflecting the different roles a verse can play within the overall structure of a 
performance.

The \textbf{first group} illustrates deictic formulas with
productive open slots.
The 5-gram \textit{de fronte a unu pópulu [ADJ]} ($PMI = 27.29$, $G^2 = 286.68$, obs.~8) is attested across eight distinct performances of Masala with five adjectival completions (\textit{signore}, \textit{devotu}, \textit{gentile}, \textit{'e zente}, \textit{festosu}) and is classified as an idiolectal formula.
The 5-gram \textit{de su palcu in susu} ($PMI = 22.95$, $G^2 = 119.27$, obs.~4) is distributed across Mura and Masala in four distinct performances, with a variable left context (\textit{da chi t'agatas}, \textit{las improviso}, \textit{chi as cantadu}, \textit{da chi m'agato}) and a frozen deictic core, indicating membership in the shared traditional repertoire.
These two sequences illustrate a gradient from idiolectal formulas with a productive terminal slot to cross-poet templates with a productive left context.

The two instances grouped under the \textbf{second group} represent a functionally coherent class of \textit{personal formulas} used to manage the poet's own turn at the opening or early internal position of a strophe. Both $n$-grams are strictly monoauctorial — all attested occurrences are attributed to a single poet — and both carry a consistent pragmatic function across distinct performance events and distinct opponents.
The first verse is the 5-gram \textit{[ADV/PRON]
una cosa narrer ti cheria} ($PMI = 29.44$, $G^2 = 194.09$, obs.~5), attested across five distinct performances of Sozu from 1958 to the early 1980s with three onset realizations (\textit{ma}, \textit{deo}, \textit{solu}).
The five-word core is lexically frozen while the initial position accommodates metrically interchangeable elements.
The formula marks a discourse-level shift to direct personal address and is exclusive to Sozu across the corpus, constituting evidence of idiolectal formulaic 
stability over a long temporal span.
The second verse, \textit{su chi ses nelzende pone cura} 
(transl.\ \textit{pay attention to what you are weaving}; 
$n = 6$, obs.\ $= 3$, $PMI = 42.72$, $G^2 = 171.66$), is 
exclusively associated with Piredda and distributed across 
three distinct performance events (1976 and twice in 1980) 
involving three distinct opponents, suggesting idiolectal 
stability over at least a four-year span.

\textbf{The third group} collects instances of dialogic mirroring, attested here by two examples.
The 8-gram \textit{a morrer de piumu est un'onore} ($PMI = 46.62$, $G^2 = 125.25$, obs.~2) occurs in two consecutive strophes of the same performance (see Table~\ref{tab:piumu} in the Appendix): Piras deploys it as the closing verse of strophe~\#62 and Sozu opens his immediate response (strophe~\#63) with that same line, recontextualizing the formula to dismantle the argument it had supported. Where Piras invokes dying by bullet as an honor applicable to the bandit, Sozu accepts the formula but redirects it to argue that such honor is not indiscriminate.
The second case, the 7-gram \textit{chi si l'at mandigada sa rustia} (strophes \#8--\#9, Masala $\to$ Mura, $PMI = 45.50$, $G^2 = 122.15$, obs.~2 see; Table~\ref{tab:rustia} 
in the Appendix)), follows the same structural logic. Masala introduces the verse as a closing statement on the poor harvest and Mura opens the following strophe by echoing it verbatim before expanding the causal frame, adding the \textit{peronòspera} (downy mildew) as a further agent of destruction.
Both cases instantiate verbatim $n$-gram repetition across a turn boundary, yet in functionally distinct ways: argumentative inversion, in which the repeated formula is turned against its original claim and elaborative extension, in which it is accepted and expanded with additional causal content.  

The \textbf{fourth functional group} collects sequences that instantiate the same pragmatic act --- asserting the opponent's failure to understand --- at different levels of lexical fixity.
The 8-gram \textit{e cando mai chi non b'as cumpresu}(Piredda, $PMI = 45.80$, $G^2 = 122.98$, obs.~2) and \textit{e cando mai chi non l'as cumpresa} (Sozu, $PMI = 45.57$, $G^2 = 122.34$, obs.~2) are near-minimal pairs sharing the invariant core \textit{cando mai chi non [PRON] as cumpress-}, differing only in clitic form and participial gender inflection. Their attribution to two distinct poets indicates a cross-poet template.
The 5-gram \textit{est chi non l'as [V-PART]} ($PMI = 16.71$, $G^2 = 63.48$, obs.~3), distributed across Mura and Sale with completions \textit{cumpresa} and \textit{iscritu}, represents a more abstract level: a productive syntactic frame with an open participial slot.

\section{Discussion}

This study aimed to address two interconnected objectives: the construction of a structured digital corpus of Sardinian extemporaneous poetry suitable for computational analysis and the empirical investigation of formulaic behavior within this tradition.

On the first front, \textbf{A Bolu} represents, to our 
knowledge, the first resource of its kind for the 
\textit{cantada logudoresa}. Its hierarchical architecture demonstrates that the fragmentary and discontinuous nature of oral tradition heritage can be transformed into a structured and computationally accessible dataset without sacrificing contextual richness.

Lexical and temporal analyses provide descriptive baselines that characterize the individual stylistic profiles of the eight poets. The shared lexical core that emerges from the aggregate count of types, combined with individual variation in hapax legomena and performance tenses, suggests that the poets operate within a common expressive framework while maintaining recognizable personal signatures. These findings provide the necessary empirical grounding for the more  theoretically significant question of formularity: the observed $n$-gram recurrences are not an artifact of lexical poverty but reflect genuine formulaic preference within a rich and varied individual repertoire.

On this front, the $n$-gram analysis yields the most significant results. The recurrence of higher-order $n$ grams across distinct competitions and the positional coherence of reused lines suggest the existence of a formulaic layer in the compositional process of the \textit{cantada logudoresa}, consistent with the \textbf{Oral-Formulaic Theory} developed by Parry and Lord~\cite{parry1987making, lord1995singer}. Originally conceived to account for the compositional mechanisms of Homeric epic poetry, the theory posits that 
oral poets rely on a shared formulaic repertoire to navigate complex metrical constraints under real-time cognitive pressure. In the context of the \textit{cantada logudoresa}, such formulas serve a dual function: they operate as mnemonic anchors and as structural building blocks, enabling the improvising poet to construct metrically and thematically coherent octaves while responding spontaneously to a rival.

The evidence further points to a layered architecture of formulaic competence, operating simultaneously at the collective and individual level. At the collective level, cross-poet templates indicate membership in a shared traditional repertoire. At the individual level, strictly 
monoauctorial formulas --- stable across different opponents and performance events spanning years or decades ---suggest that formulaic competence also has a deeply personal dimension, individually cultivated over long stretches of a poet's career, in line with the idiolectal dimension of oral-formulaic composition discussed in Foley~\cite{foley1988}.

An interesting result is the interactional dimension of this formularity. The reuse of a rival's line as the opening of one's own strophe suggests that formulas function not only as individual mnemonic devices, but also as competitive rhetorical strategies --- a dimension that the classical Oral-Formulaic Theory, developed primarily in the context of monologic epic composition, does not explicitly account for. 
Two distinct modes are attested: argumentative inversion, in which the repeated formula is turned against its original claim, elaborative extension, in which it is accepted and expanded with additional content. This finding may point to a distinctive feature of agonistic improvisational traditions more broadly.

Overall, the data are consistent with the plausibility of the oral-formulaic hypothesis applied to this tradition. 
However, given the exploratory nature of the analysis, these results should be interpreted as stimuli for further investigation rather than as a formal test of the theory. 
Larger datasets and methods capable of identifying near-variant formulas beyond exact string matching will be needed to move from suggestive evidence to robust empirical validation.

\section{Conclusion and Future Works}

This study has introduced \textbf{A Bolu}, the first structured digital corpus of Sardinian extemporaneous poetry designed for computational analysis, and has presented a preliminary investigation of lexical, temporal, and formulaic dimensions of the \textit{cantada logudoresa}. The results provide preliminary empirical support for the oral-formulaic hypothesis within this tradition and reveal an interactional dimension of formulaic reuse that warrants further theoretical attention.

Several directions for future work follow naturally from the limitations of the present study. The most pressing priority is the expansion of the corpus itself: incorporating new poetic sessions and a broader range of poets would substantially increase the statistical power of the analysis and improve the generalizability of the findings beyond the eight performers currently represented. A larger and more diverse dataset would also allow for more reliable cross-poet comparisons, which are currently constrained by the pronounced imbalance in subcorpus sizes.

A particularly promising avenue is the integration of the audio dimension. The current corpus is exclusively text-based, which means that relevant prosodic and musical features of the performance --- including melodic contour, vowel prolongation, and rhythmic patterning --- remain outside the scope of the analysis. The inclusion of aligned audio recordings would not only allow for a more accurate interpretation of execution time data, but would also open the way for acoustic and phonetic analyses capable of capturing aspects of improvisational competence that are invisible at the textual level.

Finally, the development of more sophisticated metrics for evaluating stanza complexity represents a critical methodological challenge for future work. The descriptive indices used here --- hapax counts, raw $n$-gram frequencies, TTR, MATTR and MTLD --- provide a useful baseline but are insufficient to capture the full richness of the stylistic, linguistic, and cognitive phenomena at play in real-time oral composition. Length-normalized diversity metrics, network-based models of inter-poet formulaic exchange, and methods for detecting near-variant formulas beyond exact string matching would all contribute to a more nuanced understanding of how individual poets navigate the tension between metrical constraint, thematic coherence, and improvisational speed. Ultimately, such developments would position the \textbf{A Bolu} corpus as a reference resource not only for the study of Sardinian oral traditions, but for the broader computational investigation of extemporaneous poetic performance as a window into the stylistic, linguistic, and cognitive dimensions of human creativity under constraint.

\section*{Acknowledgements}
The authors wish to thank the editorial staff of \textit{Làcanas} 
and the publisher \textit{Domus de Jana} for their kind availability 
and for granting permission to use their data in the construction 
of this corpus.

\newpage

\section{Bibliographical References}\label{sec:reference}

\bibliographystyle{lrec2026-natbib}
\bibliography{lrec2026-example}

\begin{thebibliography}{1}
\expandafter\ifx\csname natexlab\endcsname\relax\def\natexlab#1{#1}\fi

\bibitem[{Calderaro and Monti(2026)}]{ABolu2026}
Silvio Calderaro and Johanna Monti. 2026.
\newblock \emph{A Bolu: a Structured Dataset for the Computational Analysis of Sardinian Improvisational Poetry}.
\newblock PID \href{https://doi.org/10.5281/zenodo.19264263}{https://doi.org/10.5281/zenodo.19264263}.

\end{thebibliography}


\begin{thebibliography}{18}
\expandafter\ifx\csname natexlab\endcsname\relax\def\natexlab#1{#1}\fi

\bibitem[{Angioni et~al.(2018)Angioni, Tuveri, Virdis, Lai, and Maltesi}]{angioni2018}
Manuela Angioni, Franco Tuveri, Maurizio Virdis, Laura~Lucia Lai, and Micol~Elisa Maltesi. 2018.
\newblock \href {https://doi.org/10.18653/v1/2018.gwc-1.53} {{SardaNet}: {A} linguistic resource for {Sardinian} language}.
\newblock In \emph{Proceedings of the 9th Global {WordNet} Conference}, pages 412--419, Nanyang Technological University (NTU), Singapore. Global Wordnet Association.

\bibitem[{Baayen(2001)}]{baayen2001word}
R~Harald Baayen. 2001.
\newblock \emph{Word frequency distributions}, volume~18.
\newblock Springer Science \& Business Media.

\bibitem[{Carta et~al.(2025)Carta, Chessa, Contu, Corriga, Deidda, Fenu, Frigau, Giuliani, Grassi, Manca, Marras, Mola, Mossa, Mura, Ortu, Piano, Pisano, Pisu, Podda, Pompianu, Seu, and Tiddia}]{carta2025}
Salvatore~Mario Carta, Stefano Chessa, Giulia Contu, Andrea Corriga, Andrea Deidda, Gianni Fenu, Luca Frigau, Alessandro Giuliani, Luca Grassi, Marco~Manolo Manca, Mirko Marras, Francesco Mola, Bastianino Mossa, Paola Mura, Marco Ortu, Leonardo Piano, Simone Pisano, Antonella Pisu, Alessandro~Sebastian Podda, Livio Pompianu, Sara Seu, and Simona~Giuseppina Tiddia. 2025.
\newblock \href {https://clic2025.unica.it/wp-content/uploads/2025/09/17_main_long.pdf} {A {BERT}-based approach for {Part-of-Speech} tagging in the {Sardinian} language}.
\newblock In \emph{Proceedings of the Eleventh Italian Conference on Computational Linguistics ({CLiC-it} 2025)}. CEUR-WS.

\bibitem[{Chiruzzo et~al.(2022)Chiruzzo, G{\'o}ngora, Alvarez, Gim{\'e}nez-Lugo, Ag{\"u}ero-Torales, and Rodr{\'i}guez}]{guarani2022}
Luis Chiruzzo, Santiago G{\'o}ngora, Aldo Alvarez, Gustavo Gim{\'e}nez-Lugo, Marvin Ag{\"u}ero-Torales, and Yliana Rodr{\'i}guez. 2022.
\newblock \href {https://aclanthology.org/2022.lrec-1.226} {{Jojajovai}: {A} parallel {Guarani}-{Spanish} corpus for {MT} benchmarking}.
\newblock In \emph{Proceedings of the Thirteenth Language Resources and Evaluation Conference}, pages 2098--2107, Marseille, France. European Language Resources Association.

\bibitem[{Chizzoni and Vietti(2024)}]{chizzoni2024}
Ilaria Chizzoni and Alessandro Vietti. 2024.
\newblock \href {https://ceur-ws.org/Vol-3878/25_main_long.pdf} {Towards an {ASR} system for documenting endangered languages: {A} preliminary study on {Sardinian}}.
\newblock In \emph{Proceedings of the Tenth Italian Conference on Computational Linguistics ({CLiC-it} 2024)}, volume 3878 of \emph{{CEUR} Workshop Proceedings}, Pisa, Italy. CEUR-WS.

\bibitem[{Church and Hanks(1990)}]{church1990}
Kenneth~W. Church and Patrick Hanks. 1990.
\newblock Word association norms, mutual information, and lexicography.
\newblock \emph{Computational Linguistics}, 16(1):22--29.

\bibitem[{Covington and McFall(2010)}]{covington2010cutting}
Michael~A Covington and Joe~D McFall. 2010.
\newblock Cutting the gordian knot: The moving-average type--token ratio (mattr).
\newblock \emph{Journal of quantitative linguistics}, 17(2):94--100.

\bibitem[{Dunning(1993)}]{dunning1993}
Ted Dunning. 1993.
\newblock Accurate methods for the statistics of surprise and coincidence.
\newblock \emph{Computational Linguistics}, 19(1):61--74.

\bibitem[{Foley(1988)}]{foley1988}
John~Miles Foley. 1988.
\newblock \emph{The Theory of Oral Composition: History and Methodology}.
\newblock Indiana University Press, Bloomington, IN.

\bibitem[{Grobol and Jouitteau(2024)}]{arbres2024}
Lo{\"i}c Grobol and M{\'e}lanie Jouitteau. 2024.
\newblock \href {https://aclanthology.org/2024.lrec-main.130} {{ARBRES} {Kenstur}: {A} {Breton}-{French} parallel corpus rooted in field linguistics}.
\newblock In \emph{Proceedings of the Fourteenth Language Resources and Evaluation Conference}, Torino, Italy. European Language Resources Association.

\bibitem[{Lord(1995)}]{lord1995singer}
Albert~Bates Lord. 1995.
\newblock \emph{The singer resumes the tale}.
\newblock Cornell University Press.

\bibitem[{McCarthy and Jarvis(2010)}]{mccarthy2010mtld}
Philip~M McCarthy and Scott Jarvis. 2010.
\newblock Mtld, vocd-d, and hd-d: A validation study of sophisticated approaches to lexical diversity assessment.
\newblock \emph{Behavior research methods}, 42(2):381--392.

\bibitem[{Ong(1982)}]{ong1982orality}
Walter~J. Ong. 1982.
\newblock \emph{Orality and Literacy: The Technologizing of the Word}.
\newblock Routledge, London.

\bibitem[{Parry and Parry(1987)}]{parry1987making}
Milman Parry and Adam Parry. 1987.
\newblock \emph{The making of Homeric verse: The collected papers of Milman Parry}.
\newblock Oxford University Press.

\bibitem[{Ramponi(2024)}]{ramponi2022}
Alan Ramponi. 2024.
\newblock \href {https://doi.org/10.1162/tacl_a_00631} {Language varieties of {Italy}: Technology challenges and opportunities}.
\newblock \emph{Transactions of the Association for Computational Linguistics}, 12:19--38.
\newblock First appeared as arXiv preprint arXiv:2209.09757, September 2022.

\bibitem[{{Redazione Làcanas}(2014)}]{lacanas_poetasabolu}
{Redazione Làcanas}. 2014.
\newblock Poetas a bolu.
\newblock Rubrica online, \textit{Làcanas -- Rivista bilingue delle identità}.
\newblock Accessed: 2025. Available at: \url{https://www.lacanas.it/rubrica/poetas-a-bolu/}.

\bibitem[{Virdis(1988)}]{virdis1988}
Maurizio Virdis. 1988.
\newblock Sardisch: {Areallinguistik}.
\newblock In G{\"u}nter Holtus, Michael Metzeltin, and Christian Schmitt, editors, \emph{Lexikon der Romanistischen Linguistik ({LRL})}, volume~4, pages 897--913. Max Niemeyer Verlag, T{\"u}bingen.

\bibitem[{Zumthor(1997)}]{zumthor1997construction}
Paul Zumthor. 1997.
\newblock The construction of orality.
\newblock \emph{Poetry in Speech: Orality and Homeric Discourse}, page~7.

\end{thebibliography}

\section{Language Resource References}
\label{lr:ref}
\bibliographystylelanguageresource{lrec2026-natbib}
\bibliographylanguageresource{languageresource}


\clearpage
\appendix
\onecolumn

\section*{Appendix A. Dialogic Mirroring: Full Stanzas and English Translations}
\label{sec:appendix_stanzas}

\begin{center}
{\small\sffamily
\setlength{\tabcolsep}{5pt}
\renewcommand{\arraystretch}{1.1}
\begin{tabularx}{\textwidth}{@{} X | X @{}}
\toprule
\multicolumn{2}{c}{\textbf{Original Sardinian Text}} \\
\midrule
\textbf{Stanza \#62 (Piras)} & \textbf{Stanza \#63 (Sozu)} \\
\midrule
Ma cussos una fama tenen totu &
    \textbf{A morrer de piumu est un'onore} \\
e sempre de sa zente sun a galla. &
    ma cuss'onore non deghet a totu \\
A tie fatu t'an sa facia gialla &
    prite disparidade no as connotu \\
proite su valore nde as connotu. &
    o lis pones su propiu valore \\
Nara proite su Milite Ignotu &
    chi a fiancu 'e Marras e Pintore \\
no fut bandidu e moltu l'an a balla. &
    mi cheres ponner su Milite Ignotu \\
Est moltu che bandidu e gherradore: &
    ch'in su Vitorianu che at sa losa: \\
\textbf{a morrer de piumu est un'onore.} &
    creo no siat sa matessi cosa. \\
\midrule\midrule
\multicolumn{2}{c}{\textbf{English Translation}} \\
\midrule
\textit{But they all have a reputation} &
    \textbf{\textit{To die by lead is an honor}} \\
\textit{and are always held high by the people.} &
    \textit{but that honor does not suit everyone} \\
\textit{To you, they made your face turn pale} &
    \textit{because you have seen no disparity} \\
\textit{because you have known their value.} &
    \textit{or you give them the same value} \\
\textit{Tell me why the Unknown Soldier} &
    \textit{when, alongside Marras and Pintore,} \\
\textit{was not a bandit, yet he was killed by a bullet.} &
    \textit{you want to place the Unknown Soldier} \\
\textit{He died like a bandit and a warrior:} &
    \textit{who has his tomb in the Vittoriano:} \\
\textbf{\textit{to die by lead is an honor.}} &
    \textit{I believe it is not the same thing.} \\
\bottomrule
\end{tabularx}
}
\end{center}
\captionof{table}{Comparison of stanzas for the \textit{piumu} occurrence.}
\label{tab:piumu}
 
\vspace{20pt}
 
\begin{center}
{\small\sffamily
\setlength{\tabcolsep}{5pt}
\renewcommand{\arraystretch}{1.1}
\begin{tabularx}{\textwidth}{@{} X | X @{}}
\toprule
\multicolumn{2}{c}{\textbf{Original Sardinian Text}} \\
\midrule
\textbf{Stanza \#8 (Masala)} & \textbf{Stanza \#9 (Mura)} \\
\midrule
S'annu passadu s'annada fit mala, &
    \textbf{Chi si l'at mandigada sa rustia} \\
canta sicagna e canta caristia! &
    tue afirmadu as che poesianu. \\
In tota sa Sardigna de ua ebbia &
    No fit s'annu passadu cuss'ebbia, \\
si nd'at salvadu solu calchi iscala: &
    bistadu est su tribagliu totu invanu \\
fit cosa in generale e in-d-ogni ala &
    e ocannu ca est fritu su 'eranu \\
\textbf{chi si l'at mandigada sa rustia.} &
    sa 'ide l'at distruta sa 'iddia: \\
Ma bell'e gai s'oju nd'at annotu &
    s'annu passadu sende a bide pr\`ospera \\
chi su tantu 'e su 'inu ch'est etotu. &
    mandigada si l'at sa peron\`ospera. \\
\midrule\midrule
\multicolumn{2}{c}{\textbf{English Translation}} \\
\midrule
\textit{Last year the season was bad,} &
    \textbf{\textit{That the frost has eaten it all up}} \\
\textit{so much drought and so much famine!} &
    \textit{you have stated like a poet.} \\
\textit{In all of Sardinia, only a few clusters} &
    \textit{It wasn't just like that last year,} \\
\textit{of grapes were saved here and there:} &
    \textit{all the labor was in vain} \\
\textit{it was a general thing in every part} &
    \textit{and this year, because the spring is cold,} \\
\textbf{\textit{that the frost has eaten it all up.}} &
    \textit{the frost has destroyed the vine:} \\
\textit{But even so, the eye notices} &
    \textit{last year, while the vine was prosperous,} \\
\textit{that the amount of wine is still the same.} &
    \textit{the downy mildew ate it up.} \\
\bottomrule
\end{tabularx}
}
\end{center}
\captionof{table}{Comparison of stanzas for the \textit{rustia} occurrence.}
\label{tab:rustia}

\end{document}